\documentclass{article} 
\usepackage{iclr2026_conference,times}

\usepackage{amsmath,amsfonts,bm}

\def\eqref#1{equation~\ref{#1}}

\def\1{\bm{1}}

\DeclareMathAlphabet{\mathsfit}{\encodingdefault}{\sfdefault}{m}{sl}
\SetMathAlphabet{\mathsfit}{bold}{\encodingdefault}{\sfdefault}{bx}{n}

\usepackage{hyperref}
\usepackage{url}

\usepackage{graphicx}
\usepackage{textcomp}
\usepackage{booktabs} 
\usepackage{siunitx}
\usepackage{xspace,url}
\usepackage{color}
\usepackage{colortbl}
\usepackage{multirow}
\usepackage{multicol}
\usepackage{makecell}
\usepackage{tabularx}
\usepackage{listings}
\usepackage{graphicx}
\usepackage{caption}
\usepackage{enumitem}
\usepackage{adjustbox}
\usepackage{subcaption}
\usepackage{listings}
\usepackage{xcolor}
\usepackage{soul}
\usepackage{url}
\usepackage[utf8]{inputenc}
\usepackage{caption}
\usepackage{tikz}
\usepackage{amsmath,amsfonts, bm}
\usepackage{algorithm}
\usepackage{algorithmicx}
\usepackage[noend]{algpseudocode}
\usepackage{mdframed}

\title{Joint Agent Memory and Exploration Learning via Novelty Signals}

\iclrfinalcopy 
\makeatletter
\def\@oddhead{}
\def\@evenhead{}
\makeatother

\author{Shizuo Tian\textsuperscript{1}, Xiaohong Weng\textsuperscript{2}, Rui Kong\textsuperscript{3}, Yuxuan Chen\textsuperscript{1}, Guohong Liu\textsuperscript{1}, Yuebing Song\textsuperscript{4}, \\
\textbf{Jiacheng Liu\textsuperscript{5}, Yuchen Li\textsuperscript{3}, Dawei Yin\textsuperscript{3}, Ting Cao\textsuperscript{1}, Yunxin Liu\textsuperscript{1}, Yuanchun Li\textsuperscript{1}\thanks{Corresponding Author: Yuanchun Li (liyuanchun@air.tsinghua.edu.cn)}}\\
\textsuperscript{1}Tsinghua University,
\textsuperscript{2}Sun Yat-sen University, 
\textsuperscript{3}Baidu Inc.,
\textsuperscript{4}Tongji University, \\
\textsuperscript{5}Peking University}

\newcommand{\ours}{JAMEL\xspace}

\begin{document}

\maketitle

\begin{abstract}
In open-ended environments, exploration is fundamental for autonomous agents, yet current language model agents struggle with this. Effective exploration requires memory, but retaining raw interaction histories is computationally expensive over long trajectories. While latent memory offers a solution to compress interaction histories, its training lacks reliable supervisory signals. We introduce \textbf{J}oint \textbf{A}gent \textbf{M}emory and \textbf{E}xploration \textbf{L}earning (\textbf{JAMEL}), a framework that trains agentic memory and exploration policy together through novelty-driven interaction. We observe that memory and exploration form a mutually dependent loop: sustained exploration requires memory to distinguish exhausted behaviors from unseen ones, while novelty-seeking interaction provides the supervision needed to make memory useful for future exploration. By utilizing deterministic and persistent novelty signals such as code coverage in the GUI domain, we provide natural, annotation-free supervision for the memory module. Empirical evaluations demonstrate that \ours successfully generalizes to unseen environments. Its exploration capability outperforms open-weight baselines and rivals the exploration depth of a closed-source model while reducing token consumption. Our code and model are open-sourced at \url{https://github.com/MobileLLM/JAMEL}.
\end{abstract}

\section{Introduction}
\label{sec:intro}

Exploration is a fundamental capability for intelligent agents operating in open-ended environments, where extrinsic rewards are extremely sparse or absent~\citep{ICM}. Recent work has shown that current LLM-based agents lack this capability: even when agents encounter unexpected but task-relevant information during interaction, they fail to act on it in the majority of cases~\citep{englaender2026ignore}, a deficit rooted in how agents are trained rather than in inference-time configuration.

Effective exploration requires memory. In a partially observable environment, an agent cannot decide whether an action is worth trying unless it knows which states, interface regions, or behavioral consequences have already been observed. The most direct solution is to retain the full interaction history, but this becomes computationally expensive as trajectories grow longer. Agent memory is therefore needed to compress long histories into a more compact form. Agent memory has received growing attention~\citep{zhang2024survey}, and latent memory in particular has been studied for its efficiency in compressing interaction history into a vector prefix~\citep{nextmem2026}. However, the difficulty is that agent memory lacks reliable step-level supervision: we usually do not know what each memory state should encode, or how the policy should use them in future decisions. This problem becomes more severe over long trajectories, where ineffective memory causes the agent to revisit exhausted behaviors and make poor decisions. Explicit textual memories are interpretable: their contents can be inspected, revised, or heuristically filtered when the agent fails. Latent memory is usually more efficient but much harder to supervise, because its compressed vectors lack human-readable semantics. As a result, standard task demonstrations provide no clear step-level target for what the memory should encode or how the policy should use it to avoid repetition and discover new states.

We observe that exploration and memory are mutually dependent. Memory enables the agent to avoid repetition and discover unexplored behaviors, while exploration exposes which historical information is useful for future decisions. Based on this observation, exploration itself can provide supervision for memory. Novelty is awarded only when the agent reaches behavior not covered by its own history, maximizing this reward forces the memory-conditioned policy to encode and use what has already been tried.  This alignment also creates a natural curriculum: as familiar interactions are exhausted, only deeper multi-step sequences continue to yield novelty reward, driving the policy and memory to improve together without explicit curriculum design.  In some environments, a suitable novelty signal is available without annotation effort.  In the GUI domain, code coverage provides a natural proxy: any software application can be instrumented to report which code paths have been executed, yielding a deterministic and persistent measure of behavioral novelty.  In embodied environments, analogous signals arise from discovering new states or objects encountered during navigation or manipulation.

We introduce \textbf{J}oint \textbf{A}gent \textbf{M}emory and \textbf{E}xploration \textbf{L}earning (\textbf{JAMEL}) to instantiate this idea, and our contributions are as follows: 
(1) We design a latent memory architecture that compresses historical information into memory tokens, substantially
reducing the computational overhead of processing long interaction
histories.
(2) We build a data collection pipeline for training \ours's exploration
ability via rejection fine-tuning, collecting 24k training samples
across 86 web applications in the GUI domain.
(3) We show that \ours generalizes exploration to unseen applications,
outperforming existing agent memory baselines on 10 held-out apps.
 
\section{Related Work}
\label{sec:related}

\paragraph{Agent Memory}
Efficiently compressing and organizing long-term interaction histories is fundamental to enabling autonomous agents to learn continuously. Early approaches rely on fixed context windows~\citep{beltagy2020longformerlongdocumenttransformer} or external RAG retrieval~\citep{asai2023selfraglearningretrievegenerate}, which scale poorly with interaction length. In contrast, prompt tuning methods~\citep{liu2022ptuningv2prompttuning,liu2023gptunderstands} introduce trainable soft prefixes for parameter-efficient adaptation. To model longer dependencies, recurrent memory transformers~\citep{bulatov2022recurrentmemorytransformer} propagate summary states across segments, while recent advances integrate explicit or procedural memory tokens directly into transformer layers. For instance, MemoryLLM and M+~\citep{wang2024memoryllmselfupdatablelargelanguage,wang2025mextendingmemoryllmscalable} maintain layer-wise latent memory pools with controlled forgetting, and TokMem~\citep{wu2026tokmemonetokenproceduralmemory} tokenizes procedural skills for continual adaptation. Token Memory~\citep{TokenMemoryTransformer} further stores contextual information to guide generation. However, these mechanisms typically decouple memory management from the agent's learning objective, relying on static buffers or heuristic retrieval rules. 
Our approach integrates memory compression directly into the exploration loop, enabling the agent to autonomously consolidate high-value trajectories without external annotation.

\paragraph{Exploration Policy}
Efficient exploration remains a core challenge in sparse-reward environments where agents must discover valuable states without explicit supervision. Classical approaches propose intrinsic motivation signals like ICM or RND~\citep{ICM, burda2018explorationrandomnetworkdistillation} to encourage novelty through prediction or random network errors. To accelerate search under sparse rewards, archive-based methods like Go-Explore~\citep{ecoffet2021goexplorenewapproachhardexploration} maintain frontier states for targeted exploration, while variants such as IGE~\citep{lu2025intelligentgoexplorestandingshoulders} incorporate LLM-based similarity judgments, XTX~\citep{tuyls2022multistageepisodiccontrolstrategic} combines imitation learning with curiosity, and GLoW~\citep{kim2025dualscaleworldmodelsllm} leverages dual-scale world models for maintaining high-value discoveries and learning from local trial-and-error. Tree-search alternatives like MC-LAVE and MC-DML~\citep{MC-LAVE, shi2025montecarloplanninglarge} further remove the dependency on state rollback. In the GUI domain, GUI-Xplore constructs exploration videos and hierarchical downstream tasks to improve GUI agents' cross-application and cross-task generalization \citep{sun2025guixploreempoweringgeneralizablegui}, while LLM-Explorer shows that compact knowledge maintained by LLMs can guide efficient app exploration with much lower cost than step-by-step LLM action generation \citep{zhao2025llmexplorerefficientaffordablellmbased}. Our work follows this direction but focuses on a different question: how can novelty signals supervise latent agent memory so that exploration-derived knowledge can be encoded into compact memory tokens and used by the policy for future exploration?

\section{Methodology}
\label{sec:method}

\subsection{Exploration Problem}
\label{sec:method:preliminaries}

We model the exploration problem as a finite-horizon partially observable Markov decision
process,
$\mathcal{P}=(\mathcal{S},\mathcal{A},\mathcal{O},P,\Omega,\rho_0,H)$.
At step $t$, the environment has hidden state $s_t \in \mathcal{S}$, emits an
observation $o_t \sim \Omega(\cdot \mid s_t)$, receives an action
$a_t \in \mathcal{A}$, and transitions according to
$s_{t+1} \sim P(\cdot \mid s_t,a_t)$.
The agent observes the current observation and the previous interaction history
\begin{equation}
  \mathcal{H}_{<t} = ((o_1,a_1), \ldots, (o_{t-1},a_{t-1})),
  \label{eq:history}
\end{equation}
and samples actions from $\pi(\cdot \mid o_t,\mathcal{H}_{<t})$.
Exploration has no task-specific goal state.  Instead, the objective is to expose
behavior that has not appeared earlier in the same session.  We formalize this
with a novelty score
\begin{equation}
  r_t = \mathrm{Novelty}(s_t,\mathcal{H}_{<t}) \in \mathbb{R}_{\ge 0},
  \label{eq:novelty_score}
\end{equation}
where larger values indicate that the current state is less familiar given the
visited history.  The novelty function is abstract and can be instantiated in
different ways.  ICM~\citep{ICM}, for example, uses world-model
prediction error, while \ours uses rule-based signals derived
from code coverage.
The exploration objective is
\begin{equation}
  J(\pi) =
  \mathbb{E}_{\tau \sim (\pi,\mathcal{P})}
  \left[\sum_{t=1}^{H} r_t\right].
  \label{eq:exploration_objective}
\end{equation}
Under partial observability, the agent must infer from history which behavior
units have already been exhausted.

\subsection{Model Architecture of \ours}
\label{sec:method:arch}

\begin{figure}[t]
  \centering
  \includegraphics[width=\linewidth]{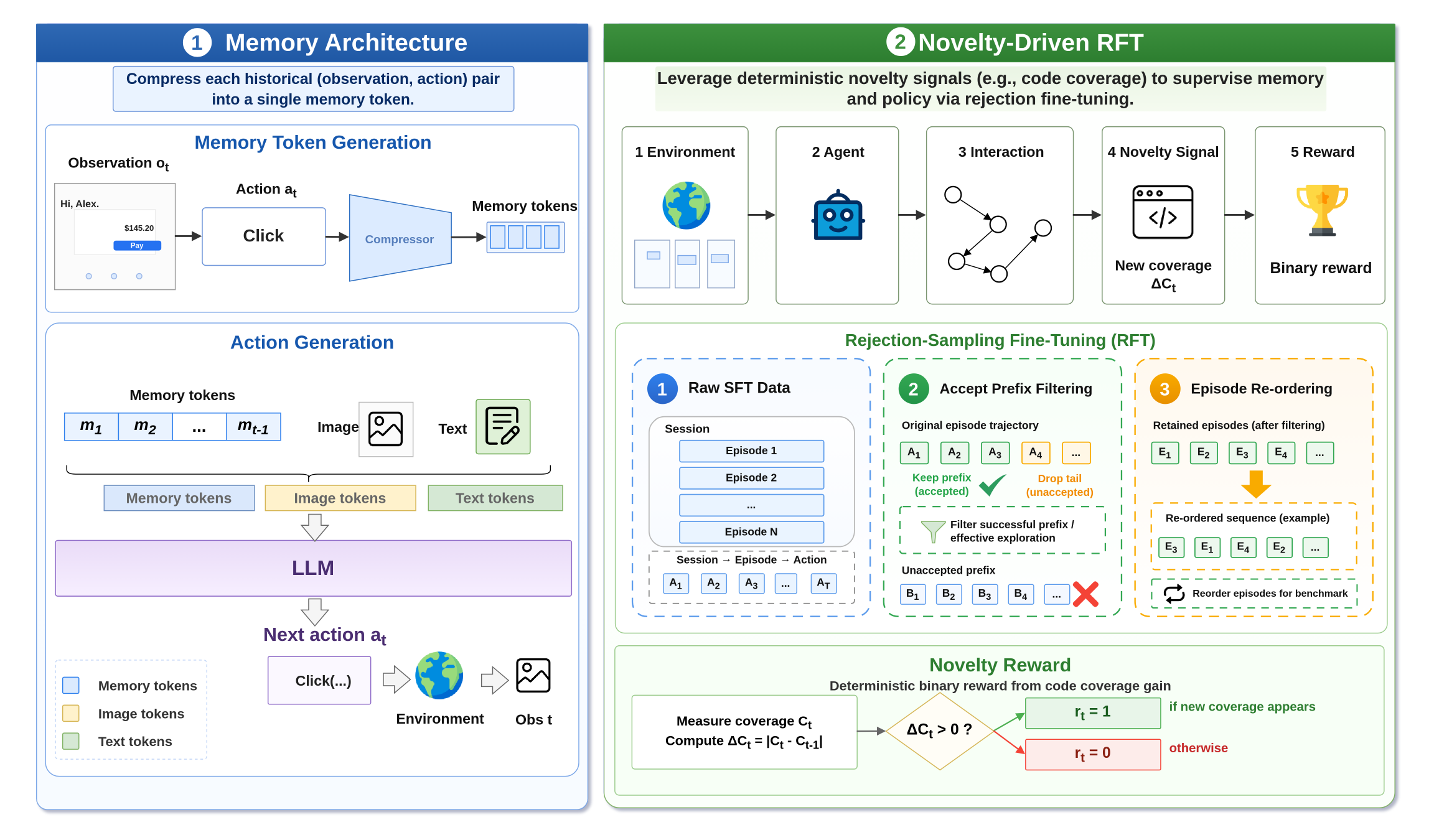}
  \caption{
    \textbf{Architecture of \ours.}
  }
  \label{fig:architecture}
\end{figure}

Let the agent's interaction history up to step \(t\) be
\(H_{<t}=\{(o_1,a_1),\ldots,(o_{t-1},a_{t-1})\}\).
For each historical step, we first serialize the observation-action pair into an input sequence
\[
x_i = \mathrm{Format}(o_i,a_i).
\]
A frozen vision-language model \(F_\phi\) is then used as the history compressor. We feed \(x_i\) into \(F_\phi\) and take the final-layer hidden state of the end-of-sequence token as a compact representation of this step:
\[
h_i = F_\phi(x_i)_{\mathrm{EOS}} \in \mathbb{R}^{d_c}.
\]
This EOS representation serves as a single latent memory token for the corresponding historical interaction. In this way, each potentially long observation-action pair is compressed into one vector, while the compressor itself remains fixed.

The memory state at time $t$ is the sequence of all the memory tokens:
\begin{equation}
  \mathbf{M}_t = [\mathbf{h}_{1}, \ldots, \mathbf{h}_{t-1}]
               \in \mathbb{R}^{(t-1) \times d_c}.
  \label{eq:memory}
\end{equation}
A learned linear aligner $\mathbf{W} \in \mathbb{R}^{d_c \times d_{\mathrm{LM}}}$
projects the memory tokens into the policy's embedding space.
The projected memory is prepended to the input embedding sequence,
\begin{equation}
  \mathbf{e}_t = [\mathbf{M}_t \mathbf{W} \;|\; \mathrm{Embed}(o_t)],
  \label{eq:prefix}
\end{equation}
and the action is then sampled as $a_t \sim \pi(\cdot \mid \mathbf{e}_t)$.


\subsection{Novelty-Based Intrinsic Reward}
\label{sec:method:reward}

Given the agent's history $\mathcal{H}_{<t}$, we define the intrinsic novelty reward at
step $t$ as a binary signal indicating whether the action produced genuinely new
experience:
\begin{equation}
  r_t = \mathbf{1}\bigl[\mathrm{Novelty}(o_t, a_t, \mathcal{H}_{<t}) > 0\bigr],
  \label{eq:reward}
\end{equation}
where $\mathrm{Novelty}$ measures how much unexplored behaviors the current
step discovers relative to the accumulated history.
The novelty measure should be persistent: once a state has
been visited, it must never register as novel again, otherwise the agent can cycle
through a small set of states and accumulate reward without genuine exploration.

In the GUI domain, code coverage provides a natural and deterministic proxy for
novelty.
Any software application can be instrumented to report which code paths have been
executed; a step is novel if and only if it triggers at least one previously
unexecuted path.
Concretely, we define the cumulative coverage score at step $t$ as
\begin{equation}
  \mathcal{C}(t) = \mathrm{cov}_{\mathrm{lines}}(t)
                 + \mathrm{cov}_{\mathrm{branches}}(t)
                 + \mathrm{cov}_{\mathrm{statements}}(t)
                 + \mathrm{cov}_{\mathrm{functions}}(t),
  \label{eq:coverage}
\end{equation}
where each term counts the total number of coverage entities executed at least once
across all steps up to and including $t$.
The intrinsic reward then simplifies to
\begin{equation}
  r_t = \mathbf{1}\bigl[\mathcal{C}(t) > \mathcal{C}(t-1)\bigr].
  \label{eq:covreward}
\end{equation}

We collect coverage data via the V8 JavaScript engine's
coverage reports and compute $\mathcal{C}(t)$ using the Istanbul
reporter~\citep{istanbul}.
Steps with invalid actions or execution errors receive $r_t = 0$.
The coverage baseline is maintained during exploration and is not reset when the
browser returns to the application's start page, so already-explored code paths
yield no reward in subsequent episodes.



Training \ours requires a dataset of exploration trajectories labeled with the
intrinsic reward defined above.
We collect this data by deploying a general-purpose LLM to explore each target
application in a browser environment.
The LLM is prompted to produce a chain-of-thought reasoning trace followed by a
single browser action at each step, operating with the full history in its
context window.
After each step, the coverage reward $r_t$ is computed according to~\eqref{eq:covreward}.

A session consists of multiple episodes, each starting from the application's
initial page and running for up to $N$ steps before a browser reset.
Because the coverage baseline is shared across episodes, the reward signal becomes
progressively sparser as the session advances, forming a natural curriculum.

From each episode, we construct a training prefix by selecting steps $1$ through
$t^*$, where $t^*$ is the index of the last step with $r_t > 0$.
Episodes with no positive-reward step are discarded.
This prefix selection ensures that every retained step belongs to a trajectory that
eventually produces novelty reward, providing a coherent learning signal for both
the policy and the memory.

For each retained step $t \leq t^*$, we pre-compute the memory tokens
$\mathbf{M}_t$ by running the compressor $F_\phi$ on all steps that precede $t$
in the session.
The training sample for step $t$ is the triple $(o_t,\, \mathbf{M}_t,\, a_t)$,
and the training objective is to maximize the likelihood of the action $a_t$ given
the current observation and memory.
The memory aligner $\mathbf{W}$ is updated jointly during this supervised phase.

We collect 24k training samples across 86 web applications on
ScaleWoB~\citep{liu2026scalewob} using this pipeline.

\section{Experiments}
\label{sec:experiments}

\subsection{Experimental Setup}
\label{sec:experiments:setup}

\paragraph{Benchmark.}
We evaluate on \textbf{ScaleWoB}~\citep{liu2026scalewob}, a benchmark of real
web applications for evaluating computer-use agents. ScaleWoB includes 96 apps spanning e-commerce,
social media and video (Weibo, Douyin, Zhihu, Youku, Tencent Video),
travel and logistics (Amap, Cainiao, Expedia), productivity
(Feishu/Lark, DingTalk, WPS), and a range of common apps.
We partition into 86 training apps and 10 test apps.  The evaluation of each app consists of $T = 50$ steps.
Agents interact with the browser via the BrowserGym action
space~\citep{chezelles2025browsergym,drouin2024workarena}, which covers clicks,
form fills, scrolls, navigation, etc.  At each step, the agent
receives the page accessibility tree (a11y tree) and the list of currently
interactive element identifiers as its observation; the image-based variants
additionally receive a screenshot.

\paragraph{Metrics.}
We report \textbf{cumulative coverage reward}: the total number of steps in a
session at which the agent's action caused the cumulative JavaScript coverage
score to increase.  Formally, each step contributes $r_t = 1$ if
$\mathcal{C}(t) > \mathcal{C}(t{-}1)$ and $r_t = 0$ otherwise.  A higher value
indicates that the agent triggers more distinct code paths across the session,
reflecting stronger exploration ability.

\paragraph{Baselines.}
Baselines are evaluated under the same ScaleWoB environment and 50-step budget. As shown in Table \ref{tab:main}, these include:
\begin{itemize}
    \item \textbf{ReAct-text} and \textbf{ReAct-vision}~\citep{yao2023react}: We implement
    the ReAct framework on top of the Gemini 3.1 Flash-Lite~\citep{googledeepmind2026gemini31flashlite}.  The text variant provides the
    agent with the full session trajectory as an AXTree-only text prompt, retaining all
    prior (observation, think, action, reward) tuples within a token budget of
    $\sim$1M tokens and without dropping any step records.
    The ReAct-vision variant appends a webpage screenshot of the current observation.
    \item \textbf{MAI-UI}~\citep{maiui2025}: A family of foundation GUI agents (2B, 8B, 32B,
    and 235B-A22B variants) built on Qwen3-VL.  MAI-UI employs a device/cloud
    collaboration mechanism that routes each step to either an on-device or cloud model
    based on task complexity.  The complete interaction trajectory is stored locally and
    reformatted per model tier; on execution errors, an error summary is appended to the
    history before the next decision.  We evaluate the 8B variant (MAI-UI-8B).
    \item \textbf{Mobile-Agent-v3.5}~\citep{mobileagentv35}: An end-to-end GUI automation
    framework built on GUI-Owl-1.5 (available in 2B, 4B, 8B, 32B, and 235B variants).
    Mobile-Agent-v3.5 uses a \emph{hierarchical context compression} strategy: the most
    recent steps retain full screenshots and action history, while earlier steps are
    distilled into concise action-conclusion summaries; a dedicated Notetaker module
    maintains a running log of task-critical information across steps.
    We evaluate the 8B variant (GUI-Owl-1.5-8B).
\end{itemize}

\paragraph{Implementation Details.}
We use Qwen3-VL-2B-Instruct~\citep{bai2025qwen3vltechnicalreport} with $d_c = 2048$ as the history compressor. Each (observation, action)
pair at step $t$ is compressed into a single memory token.
The memory tokens are projected into the embedding space of policy model, which is based on Qwen2.5-VL-7B-Instruct with $d_{\mathrm{LM}} = 3584$ via a learned linear aligner and prepended as a
soft prefix. \ours-9B is then fine-tuned using 24k samples collected from 86 apps. During evaluation, it is tested on 10 unseen apps to assess its generalization capability.

\subsection{Main Results}
\label{sec:experiments:main}

\begin{table}[htbp]
  \centering
  \caption{
    \textbf{Main results.}
    Cumulative coverage reward averaged over 10 test apps over 50 steps. \textbf{Bold} marks the best
    result and \underline{underline} marks the second-best result.
  }
  \label{tab:main}
  \setlength{\tabcolsep}{4pt}
  \renewcommand{\arraystretch}{1.12}
  \small
  \begin{tabular}{@{}>{\raggedright\arraybackslash}p{0.20\linewidth}
                  >{\raggedright\arraybackslash}p{0.32\linewidth}
                  >{\raggedright\arraybackslash}p{0.28\linewidth}
                  r@{}}
    \toprule
    \textbf{Method} & \textbf{Model} & \textbf{Memory}
      & \textbf{Avg. reward} \\
    \midrule
    \multicolumn{4}{@{}l}{\textit{Closed-source Models}} \\
    ReAct-text
      & Gemini 3.1 Flash-Lite
      & Full history
      & 19.9 \\
    ReAct-vision
      & Gemini 3.1 Flash-Lite
      & Full history
      & \textbf{20.9} \\
    \midrule
    \multicolumn{4}{@{}l}{\textit{Open-Source Models}} \\
    MAI-UI
      & MAI-UI-8B
      & Device/cloud routing
      & 8.4 \\
    Mobile-Agent-v3.5
      & GUI-Owl-1.5-8B
      & Sliding window + Notetaker
      & 5.9 \\
    \addlinespace[2pt]
    \midrule
    \textbf{\ours}
      & \ours-9B
      & Latent memory
      & \underline{20.7} \\
    \bottomrule
  \end{tabular}
\end{table}

As shown in Table~\ref{tab:main},  among small GUI
agents, \ours achieves the highest reward, compared with 
MAI-UI and Mobile-Agent-v3.5.  Against Gemini 3.1 Flash-Lite
ReAct baselines, \ours remains competitive: it exceeds ReAct-text 
and trails ReAct-vision by only 0.2 reward, despite using a 2B memory compressor and a 7B decoder.

\begin{figure}[h!]
  \centering
  \includegraphics[width=0.8\linewidth]{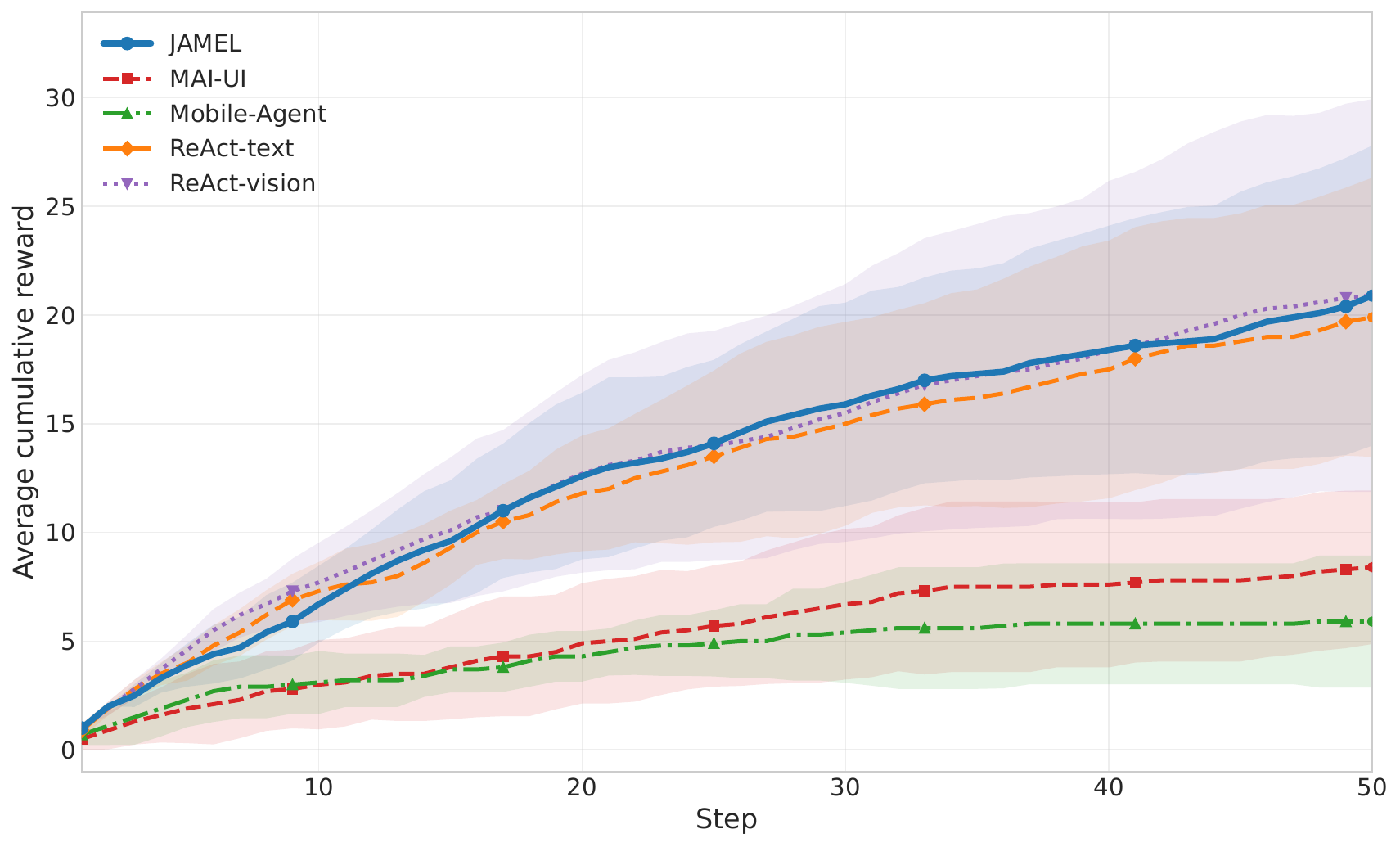}
  \caption{
    \textbf{Reward accumulation on test apps.}
    Average cumulative coverage reward across 10 test apps over a 50-step
    session.  Shaded bands denote standard error across apps.
  }
  \label{fig:reward_curve}
\end{figure}

As illustrated in Figure~\ref{fig:reward_curve}, local baselines exhibit an early plateau due to premature exploration stagnation. This stagnation occurs plausibly because these methods prune their context, inevitably discarding critical historical information as the session progresses. Cloud baselines avoid this issue by retaining the complete explicit interaction history without any pruning. \ours aligns with this comprehensive retention strategy by avoiding context truncation. Instead, \ours compresses all historical information into latent memory tokens. This mechanism allows \ours to continuously uncover new application states without stalling, maintaining a steady and continuous upward trajectory that closely tracks the performance of the cloud models, successfully achieving an exploration depth highly competitive with cloud baselines while remaining significantly more efficient.

\subsection{Analysis}
\label{sec:experiments:analysis}

\paragraph{Reward curve and the natural curriculum.}
Figure~\ref{fig:reward_curve} shows average 
cumulative reward curve for our method and baselines on the test apps.
Reward accumulates rapidly in early steps as shallow, easily-triggered interactions
are exhausted, then slows as the agent must discover deeper multi-step paths.
This sparsifying dynamic acts as a natural curriculum: the training signal becomes
progressively harder, compelling the policy to develop richer exploration strategies
over time.

\paragraph{Token efficiency.}
Table~\ref{tab:token_efficiency} reports the total input tokens consumed over
all 500 evaluation steps and the corresponding per-step average.  ReAct-text and
ReAct-vision incur the largest context cost because they retain long explicit
histories and screenshots.  Mobile-Agent-v3.5 and MAI-UI reduce this cost, but still require
2.76$\times$ and 2.81$\times$ more tokens than \ours, respectively.  Our latent
prefix keeps context compact, making exploration substantially more token
efficient.

\begin{table}[t]
  \centering
  \caption{
    \textbf{Token Consumption.}
    Input token consumption over 10 test apps with 50 steps per app. Lower is better.  \textbf{Bold} marks the best
    result.
  }
  \label{tab:token_efficiency}
  \setlength{\tabcolsep}{6pt}
  \renewcommand{\arraystretch}{1.12}
  \small
  \begin{tabular}{@{}l l r r r@{}}
    \toprule
    \textbf{Method} & \textbf{Input Tokens}
      & \textbf{Avg. per Step} & \textbf{Rel. to \ours} \\
    \midrule
    Mobile-Agent-v3.5
      & {2,931,946}
      & {5,863.9}
      & {2.76$\times$} \\
    MAI-UI
      & 2,980,061
      & 5,960.1
      & 2.81$\times$ \\
    ReAct-Text
      & 18,938,833
      & 37,877.7
      & 17.85$\times$ \\
    ReAct-Vision
      & 23,260,296
      & 46,520.6
      & 21.92$\times$ \\
    \midrule
    \textbf{\ours}
      & \textbf{1,061,103}
      & \textbf{2,122.2}
      & \textbf{1.00$\times$} \\
    \bottomrule
  \end{tabular}
\end{table}

\begin{figure}[htbp!]
  \centering
  \includegraphics[width=\columnwidth]{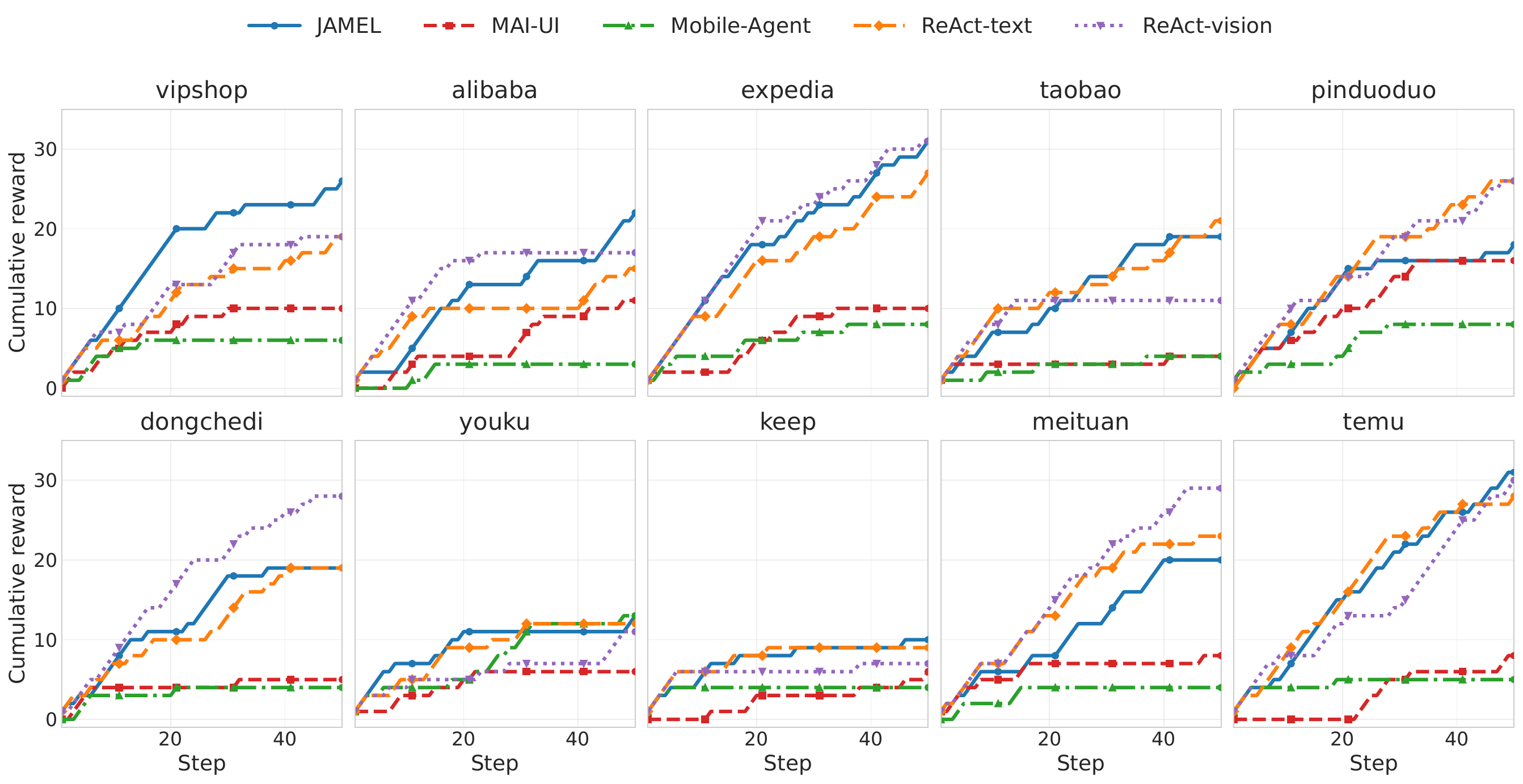}
    \caption{\textbf{Per-app reward accumulation.} Cumulative coverage reward trajectories evaluated on individual unseen applications.}
    \label{fig:reward_curve_per_app}
\end{figure}

\paragraph{Exploration Patterns.}
The per-app breakdown in Figure~\ref{fig:reward_curve_per_app} reveals distinct structural exploration patterns dictated by varying software environments. In structurally deep commerce and travel platforms such as Vipshop, Expedia, and Temu, \ours maintains continuous upward trajectories, demonstrating its capacity for sustained sequential navigation. Convers   ely, media and lifestyle applications like Youku and Keep impose a natural exploration ceiling where all methods inevitably plateau early probably due to inherently limited interactive depth. Furthermore, the stepwise reward surges observed in Alibaba and Taobao highlight the ability of \ours to escape local optima. While context pruning causes local baselines to stagnate within initial screens, the latent memory of \ours enables the policy to synthesize past interactions and transition into new application modules. However, \ours remains occasionally susceptible to entrapment. As observed in Pinduoduo, \ours experiences prolonged plateaus, suggesting that exceptionally dense interfaces can still challenge the compressed memory representation and temporarily hinder continuous state discovery.

\paragraph{Case Study.}
A detailed examination of the failure modes highlights specific interaction challenges within complex environments like Pinduoduo. The primary obstacle in such applications arises from persistent modal overlays. In these scenarios, the agent frequently attempts to interact with background elements that appear visually available but are rendered functionally unresponsive by the active foreground window.  Conversely, applications featuring straightforward graphical layouts without frequent overlay interruptions, such as Expedia, facilitate highly efficient exploration. Within these cleaner environments, \ours systematically leverages persistent structural components like bottom navigation menus to transition seamlessly between distinct functional modules.

\section{Discussion}

\paragraph{Scaling Laws of Exploration.} 
A promising direction for future research lies in exploring the scaling laws of novelty-driven memory architectures like \ours. Integrating Reinforcement Learning (RL) strategies presents a natural evolution. Because the novelty reward inherently provides a natural curriculum, it facilitates the progressive learning of complex environment operations---advancing smoothly from shallow interactions to deep, multi-step navigation. Investigating how larger model capacities, scaled-up autonomous data collection and more exploration steps can further benefit from this novelty-guided curriculum remains an open frontier for future work.

\paragraph{Memory-Conditioned Task Execution and Continual Learning.} 
Furthermore, the latent memory generated during the exploration process holds significant potential to benefit specific downstream tasks. This naturally points toward an ``explore-then-execute'' paradigm: a model first autonomously explores an unknown environment to accumulate structural memory, and subsequently relies on these exploration results to execute specific user instructions. Such an approach offers a pathway for agent self-evolution and continual learning. Developing algorithms that drive models to autonomously explore new environments and internalize new capabilities stands as a critical future direction for enabling rapid adaptation to long-tail scenarios and reducing the reliance on human-annotated trajectories.

\section{Conclusion}
We presented \ours, a framework for training agentic latent memory via novelty-driven exploration. We addressed the lack of explicit supervision for memory modules by demonstrating that persistent novelty signals, such as application code coverage in the GUI domain, can supervise agentic memory and exploration policy together by rewarding actions that use past experience to discover unexplored behaviors. Our empirical results show that \ours successfully generalizes to unseen environments, outperforming current open-weight baselines and rivaling the exploration depth of a closed-source model while reducing token consumption. While dense interfaces with persistent modal overlays can occasionally challenge the compressed representation, \ours establishes a scalable paradigm for autonomous agents in which memory and exploration are not separate modules, but mutually reinforcing capabilities learned through persistent novelty signals.

\bibliography{reference}
\bibliographystyle{iclr2026_conference}

\appendix
\end{document}